\documentclass[11pt,a4paper]{article}
\usepackage[hyperref]{swisstext2018}
\usepackage{times}
\usepackage{latexsym}

\usepackage{amsmath}
\usepackage{amssymb}
\usepackage{url}

\usepackage{booktabs}

\usepackage{array}
\newcolumntype{L}[1]{>{\raggedright\let\newline\\\arraybackslash\hspace{0pt}}m{#1}}
\newcolumntype{C}[1]{>{\centering\let\newline\\\arraybackslash\hspace{0pt}}m{#1}}
\newcolumntype{R}[1]{>{\raggedleft\let\newline\\\arraybackslash\hspace{0pt}}m{#1}}

\title{On the effectiveness of feature set augmentation using clusters of word embeddings}

\author{Georgios Balikas\\
  Kelkoo\\
  Grenoble, France\\
  {\tt georgios.balikas@kelkoogroup.com} \\\And
  Ioannis Partalas \\
  Expedia\\
  Geneva, Switzerland\\
  {\tt ipartalas@expedia.com} \\
}

\date{}

\begin{document}
\maketitle
\begin{abstract}
 Word clusters have been empirically shown to offer important performance improvements
 on various Natural Language Processing (NLP) tasks. Despite their importance, their incorporation
 in the standard pipeline of feature engineering relies more on a trial-and-error
 procedure where one evaluates several hyper-parameters, like the number of clusters to be used.
 In order to better understand the role of such features in NLP tasks we
 perform a systematic empirical evaluation on three tasks, that of named entity recognition,
    fine grained sentiment classification and fine grained sentiment quantification. 
\end{abstract}
\section{Introduction}
Many research attempts have proposed novel features that improve the performance of learning algorithms in particular tasks. Such features are often motivated by domain knowledge or manual labor. Although useful and often state-of-the-art, adapting such solutions on NLP systems across tasks can be
tricky and time-consuming \cite{DeySK16}. 
Therefore,  simple yet general and powerful methods that perform well across several datasets are valuable \cite{turian2010word}.

An approach that has become extremely popular lately in NLP tasks, is to train word embeddings in an unsupervised way. These embeddings are dense vectors that project words or short text spans like phrases in a vector space where dimensions are supposed to capture text properties. Such embeddings can then be used either as features with off-the-shelf algorithms like Support Vector Machines, or to initialize deep learning systems \cite{Goldberg15c}. However, as shown in \cite{wang2013effect} linear
architectures perform better in high-dimensional discrete spaces compared to continuous ones. The latter is probably the main reason of the high performance of the vector space model \cite{SaltonWY75} in tasks like text classification with linear models like SVMs. Using linear algorithms, while taking advantage of the expressiveness of text embeddings is the focus of this work.

In this paper, we explore a hybrid approach, that uses text embeddings as a proxy to create features. 
Motivated by the argument that text embeddings manage to encode the semantics of text, we explore how clustering text embeddings can impact the performance of different NLP tasks. 
Although such an approach has been used in different studies during feature engineering, the selection of word vectors and the 
number of clusters remain a trial-end-error procedure. In this work we present an empirical evaluation across diverse tasks to verify whether and when such features are useful. 

Word clusters have been used as features in various tasks like Part-of-Speech tagging and NER.  Owoputi et al. \shortcite{Owoputi13} use Brown clusters \cite{Brown92} in a POS tagger showing
that this type of features carry rich lexical knowledge as they can substitute lexical resources like gazetteers.  Kiritchenko et al. \shortcite{KiritchenkoZM14} discusses their use on sentiment classification while Hee et al. \shortcite{HeeLH16} incorporate them in the task of irony detection in Twitter.  Ritter et al. \shortcite{Ritter2011} inject also
word clusters in a NER tagger. While these works show that word clusters are beneficial no clear guidelines can be concluded of how and when to use them.

In this work, we empirically demonstrate that using different types of embeddings on three NLP tasks with twitter data we manage to achieve better or near to the state-of-the art performance on three NLP tasks: (i) Named Entity Recognition (NER) segmentation, (ii) NER classification, (iii) fine-grained sentiment analysis and (iv) fine-grained sentiment quantification. For each of the three tasks, we achieve higher performance than without using features which indicates the effectiveness of the cluster membership features. Importantly, our evaluation compared to previous work \cite{guo2014revisiting} who focus on old and well studied datasets uses recent and challenging datasets composed by tweets. The obtained results across
all the tasks permits us to reveal important aspects of the use of word clusters and therefore provide guidelines. Although our obtained scores are state-of-the-art, our analysis reveals that the performance in such tasks is far from perfect and, hence, identifies that there is still much space for improvement and future work.
\section{Word Clusters}
Word embeddings associate words with dense, low-dimensional vectors. Recently, several models have been proposed in order to obtain these embeddings. Among others, the skipgram (\texttt{skipgram})  model with negative sampling \cite{abs-1301-3781}, the continuous bag-of-words (\texttt{cbow}) model \cite{abs-1301-3781} and Glove (\texttt{glove}) \cite{glove} have been shown to be effective. Training those models requires no annotated data and can be done using big amounts of text. Such a model can be seen as a function $f$ that projects a word $w$ in a $D$-dimensional space: $f(w) \in \mathbb{R}^D$, where $D$ is predefined. Here, we focus on applications using data from Twitter, which pose several difficulties due to being particularly short, using creative vocabulary, abbreviations and slang.

For all the tasks in our experimental study, we use 36 millions English tweets collected between August and September 2017.
A pre-processing step has been applied to replace URLs with a placeholder and to pad punctuation.
The final vocabulary size was around 1.6 millions words.\footnote{We trained the word embeddings using the implementations released from the authors of the papers. Unless differently stated we used the default parameters.}
Additionally to the in-domain corpus we collected, we use GloVe vectors trained on Wikipedia articles
in order to investigate the impact of out-of-domain word-vectors.\footnote{The pre-trained vectors are obtained from \url{http://nlp.stanford.edu/projects/glove/}}

We cluster the embeddings with $k$-Means. The k-means clusters are initialized using ``k-means++'' as proposed in \cite{arthur2007k}, while the algorithm is run for 300 iterations. We try different values for $k\in\{100, 250, 500, 1000, 2000\}$. For each $k$, we repeat the clustering experiment with different seed initialization for 10 times and we select the clustering result that minimizes the cluster inertia. 

\section{Experimental Evaluation}
We evaluate the proposed approach for augmenting the feature space in four tasks: (i) NER segmentation,  (ii) NER classification, (iii) fine-grained sentiment classification and (iv) fine-grained sentiment quantification. 
The next sections present the evaluation settings we used. For each of the tasks, we use the designated training sets to train the learning algorithms, and we report the scores of the evaluation measures used in the respective test parts.

\subsection{Named-Entity Recognition in Twitter}
NER concerns the classification of textual segments in a predefined
set of categories, like persons, organization and locations.
We
use the data of the last competition in NER for Twitter which released as a part
of the 2nd Workshop on Noisy User-generated Text \cite{Strauss16}.
More specifically, the organizers provided annotated tweets with 10 named-entity types
(person, movie, sportsteam, product etc.) and the task comprised two sub-tasks: 1) the 
detection of entity bounds and 2) the classification of an entity into one of the 10
types. The evaluation measure for both sub-tasks is the F$_{1}$ measure.

The following is an example of a tweet which contains two named entities. Note that named entities may span several
words in the text:

\vspace{0.08in}
\fbox{
	\begin{minipage}{7cm}
		$\overbrace{\hbox{CLUB BLU}}^{Facility}$ tonite ... 90 's music .. oldskool night wiith $\underbrace{\hbox{dj finese}}_{Musicartist}$
\end{minipage}}

\paragraph{Learning algorithm}
Our model for solving the task is a learning to search approach.
More
specifically we follow \cite{Partalas16c} which has been ranked 2nd among 10 participants in the aforementioned competition \cite{Strauss16}.
The model uses handcrafted features like n-grams, part-of-speech tags, capitalization and membership in gazetteers.
The algorithm used belongs to the family of learning to search for structured prediction tasks \cite{Daume14}. These methods
decompose the problem in a search space with states, actions and policies and then learn a hypothesis controlling
a policy over the state-action space. 
The BIO encoding is used for attributing the corresponding labels to the tokens where \texttt{B-type} is used for the
first token of the entity, \texttt{I-type} for inside tokens in case of multi-term entities and \texttt{O} for
non entity tokens.

\paragraph{Results}
Tables \ref{tbl:ner1} and \ref{tbl:ner2} present the results for the different number of clusters across
the three vector models used to induce the clusters.  For all the experiments we keep the same parametrization for the learning algorithm and we present the performance of each
run on the official test set.

Regarding the segmentation task we notice that adding word clusters as features improve the performance of the best model
up to 1.1 F-score points while it boosts performance in the majority of cases. In only one case, for \texttt{glove$_{40}$} vectors,
there is a drop across all number of clusters used.

As for the number of clusters, the best results are generally obtained between 250 and 1000 classes for all word vector
models. These dimensions seem to be sufficient for the three-class sub-task that we deal with.
The different models of word vectors perform similarly and thus one cannot privilege a certain type of word vectors. 
Interestingly, the clusters learned on the Wikipedia GloVe vectors offer competitive
performance with respect to the in-domain word vectors used for the other cases
showing that one can rely to out-of-domain data for constructing such representations.

\begin{table}\scriptsize
	\begin{tabular}{lrrrrr}\toprule
		& 100 & 250 & 500 & 1000 & 2000 \\ \midrule
		No clusters & & & 55.29 & &  \\ 
		skipgram$_{40,w5}$ & 55.21 & 54.42 & 55.4 & \textbf{55.78} & 55.40 \\
		skipgram$_{100,w5}$ & 54.94 & \textbf{55.32} & 54.02 & 54.51 & 53.79 \\
		skipgram$_{100,w10}$ & 55.04 & 55.37 & \textbf{55.93} & 55.00 & 54.77 \\
		cbow$_{40,w5}$ & \textbf{55.97} & 55.22 & 55.89 & 55.61 & 54.64 \\
		cbow$_{100,w5}$ & 55.76 & \textbf{56.36} & 55.03 & 55.00 & 55.60 \\
		cbow$_{100,w10}$ & 55.72 & 55.55 & 56.08 & \textbf{56.19} & 55.46 \\
		glove$_{40,w5}$ & 53.73 & 53.93 & 53.96 & 54.58 & 55.07 \\
		glove$_{100,w5}$ & 53.91 & 55.09 & 53.11 & 54.69 & \textbf{55.81} \\
		glove$_{100,w10}$ & 55.50 & \textbf{55.84} & 55.33 & 53.73 & 55.15 \\ \midrule
		glove$_{50,wiki}$ &55.56 &55.50 &\textbf{55.92} &55.87 &53.96 \\
		glove$_{100,wiki}$ &55.54 &\textbf{56.38} &55.68 &54.74 &54.91 \\
		\bottomrule
	\end{tabular}
	\caption{Scores on F1-measure for named entities segmentation for the different
		word embeddings across different number of clusters. For each  embedding type, we show its dimension and window size. For instance, glove$_{40,w5}$ is 40-dimensional glove embeddings with window size  5.}
	\label{tbl:ner1}
\end{table}

Concerning the classification task (Table \ref{tbl:ner2}) we generally observe  a drop in the performance
of the tagger as we deal with 10 classes. This essentially corresponds to a multi-class problem
with 21 classes: one for the non-entity type and two classes for each entity type.
In this setting we notice that the best results are obtained in most cases for
higher number of classes (1000 or 2000) possibly due to a better discriminatory
power in higher dimensions. Note also, that in some cases the addition of word cluster features
does not necessarily improve the performance. Contrary, it may degrade it
as it is evident in the case of \texttt{glove$_{40,w5}$} word clusters.
Like in the case of segmentation we do not observe a word vector model
that clearly outperforms the rest. Finally, we note the same competitive performance
of the Wikipedia word clusters and notably for the \texttt{glove$_{100,wiki}$}
clusters which obtain the best F1-score.

\begin{table}\scriptsize
	\begin{tabular}{lrrrrr}\toprule
		& 100 & 250 & 500 & 1000 & 2000 \\ \midrule
		No clusters & & & 40.10 & &  \\ 
		skipgram$_{40,w5}$ & 40.07 & 39.29 & 39.98 & 40.15 & \textbf{40.70} \\
		skipgram$_{100,w5}$ & 40.03 & 39.63 & 39.42 & \textbf{40.25} & 40.06 \\
		skipgram$_{100,w10}$ & \textbf{41.19} & 39.92 & 40.04 & 40.29 & 38.54 \\
		cbow$_{40,w5}$ & 40.20 & \textbf{40.63} & 39.71 & 40.00 & 39.32 \\
		cbow$_{100,w5}$ & 40.53 & 40.21 & 39.22 & \textbf{40.61} & 40.15 \\
		cbow$_{100,w10}$ & 40.02 & \textbf{40.80} & 39.55 & 40.40 & 39.71 \\
		glove$_{40,w5}$ & 39.09 & \textbf{40.06} & 38.73 & 39.35 & 38.23 \\
		glove$_{100,w5}$ & 39.63 & 39.40& 39.23 & 39.20 & \textbf{41.42} \\
		glove$_{100,w10}$ & 39.55 & 39.98 & 39.99 & 39.53 & \textbf{40.67} \\\midrule
		glove$_{50,wiki}$ &40.40 &41.09 &40.82 &40.84 &\textbf{41.24} \\
		glove$_{100,wiki}$ &39.52 &39.78 &41.04 &\textbf{41.73} &40.39 \\ \bottomrule
	\end{tabular}
	\caption{Results in terms of F1-score for named entities classification for the different
		word clusters across different number of clusters.}
	\label{tbl:ner2}
\end{table}

\subsection{Fine-grained Sentiment Analysis}
The task of fine grained sentiment classification consists in predicting the sentiment of an input text according to a five point scale (sentiment $\in$ \{\texttt{VeryNegative, Negative, Neutral, Positive, VeryPositive}\}). We use the setting of task 4 of SemEval2016 ``Sentiment Analysis in Twitter'' and the dataset released by the organizers for subtask 4 \cite{NakovRRSS16}. 

In total, the training (resp. test)  data consist of 9,070 (resp. 20,632) tweets.

The evaluation measure selected in \cite{NakovRRSS16} for the task in the macro-averaged Mean Absolute Error (MAE$^M$). It is a measure of error, hence lower values are better. The measure's goal is to take into account the order of the classes when penalizing the decision of a classifier. For instance,  misclassifying a very negative example as very positive is a bigger mistake than classifying it as negative or neutral. Penalizing a classifier according to how far the predictions are from the true class is captured by MAE$^M$ \cite{BaccianellaES09}. Also, the advantage of using the macro- version instead of the standard version of the measure is the robustness against the class imbalance in the data. 

\textbf{Learning algorithm}
To demonstrate the efficiency of cluster membership features  we rely on the system of \cite{BalikasA16} which was ranked 1$^{st}$ among 11 participants and uses a Logistic Regression as a learning algorithm. We follow the same feature extraction steps which consist of extracting n-gram and character n-gram features, part-of-speech counts as well as sentiment scores  using standard sentiment lexicons such as the Bing Liu's \cite{HuL04} and the MPQA lexicons \cite{WilsonWH05}. For the full description, we refer the interested reader to \cite{BalikasA16}. 

\paragraph{Results} To evaluate the performance of the proposed feature augmentation technique, we present in Table \ref{tbl:sent_scores} the macro-averaged Mean Absolute Error scores for different settings on the official test set of \cite{NakovRRSS16}.  First, notice that the best score in the test data is achieved using cluster membership features, where the word embeddings are trained using the skipgram model. The achieved score improves the state-of-the art on the dataset, which to the best of our knowledge was by \cite{BalikasA16}. Also, note that the score on the test data improves for each type of embeddings used, which means that augmenting the feature space using cluster membership features helps the sentiment classification  task. 

Note, also, that using the clusters produced by the out-of-domain embeddings trained on wikipedia that were released as part of \cite{glove} performs surprisingly well. One might have expected their addition to  hurt the performance. However, their value  probably stems from the sheer amount of data used for their training as well as the relatively simple type of words (like awesome, terrible) which are discriminative for this task.
Lastly, note that in each of the settings, the best results are achieved when the number of clusters is within $\{500, 1000, 2000\}$ as in the NER tasks. Comparing the performance across the different  embeddings, one cannot claim that a particular embedding performs better. It is evident though that augmenting the feature space with feature derived using the proposed method, preferably with in-domain data, helps the classification performance and reduces MAE$^M$.

\begin{table}\scriptsize\centering\setlength{\tabcolsep}{5pt}
	\begin{tabular}{l ccccc}
		\toprule
		clusters& 100 &250 &500&1000 &2000\\
		\midrule
		no clusters & \multicolumn{5}{c}{0.721} \\
		skipgram$_{40,w5}$ & 0.712 & 0.691 & 0.699 & \textbf{0.665} & 0.720 \\
		skipgram$_{100,w5}$ & 0.69 & 0.701 & \textbf{0.684} & 0.706 & 0.689 \\
		skipgram$_{100,w10}$ & 0.715 & 0.722 & 0.700 & 0.711 & \textbf{0.681} \\

		cbow$_{40, w5}$ &  0.705 & 0.707 & 0.690 & 0.700 & \textbf{0.675}    \\
		cbow$_{100,w5}$ & 0.686 & 0.699 & 0.713 & \textbf{0.683} & 0.697 \\
		
		cbow$_{100,w10}$ & 0.712 & 0.690 & 0.707 & \textbf{0.676} & 0.703 \\
		
		glove$_{40,w5}$     &  0.680 & 0.689 & \textbf{0.679} & 0.702 & 0.706    \\
		glove$_{100,w5}$    &  0.687 & 0.670 & \textbf{0.668} & 0.678 & 0.693  \\
		glove$_{100,w10}$  &  0.696 & 0.689 & 0.693 & \textbf{0.687} & 0.703    \\\midrule
		
		glove$_{50, wiki}$& 0.712 & \textbf{0.678} & 0.710 & 0.711 & 0.694 \\
		glove$_{100,wiki}$& 0.699 & 0.701 & 0.699 & \textbf{0.689} & 0.714 \\
		
		\bottomrule
	\end{tabular}
	\caption{MAE$^M$ scores (lower is better)  for sentiment classification  across different types of word embeddings and number of clusters.  }\label{tbl:sent_scores}
\end{table}

From the results of Table \ref{tbl:sent_scores} it is clear that the addition of the cluster membership features improves the sentiment classification performance. To better understand though why these clusters help, we manually examined a sample of the words associated with the clusters. To improve the eligibility of those results we first removed the hashtags and we filter the results using an English vocabulary. In Table \ref{tbl:example_clusters} we present sample words from two of the most characteristic clusters with respect to the task of sentiment classification. Notice how words with positive and negative meanings are put in the respective clusters.

\begin{table}\scriptsize
	\begin{tabular}{|l|L{5.5cm}|}
		\hline
		Positive & hotness, melodious, congratulation, wishers, appreciable, festivity, superb, amazing, awesomeness, awaited, wishes, joyous, phenomenal, anniversaries, memorable, heartiest, excitement, birthdays, unforgettable, gorgeousness, \\\hline
		Negative & jerk, damnit, carelessly, exaggerates, egoistic, panicking, fault, problematic,  harasses, insufferable, terrible, bad, psycho, stupidest, screwed, regret, messed, messes, rest, pisses, pissed, saddest, intimidating, unpleasant, ditched, negative\\\hline
	\end{tabular}
	\caption{Sample from two clusters that were found useful for the sentiment classification. Words with positive or negative meaning are grouped together. }\label{tbl:example_clusters}
\end{table}

\subsection{Fine-Grained Sentiment Quantification}
Quantification is the problem of estimating the prevalence of a class in a dataset. 
While classification concerns assigning a category to a single instance, like labeling a tweet with the sentiment it conveys, the goal of quantification is, given a set of instances, to estimate the relative frequency of single class. Therefore, sentiment quantification tries to answer questions like ``Given a set of tweets about the new iPhone, what is the fraction of \textit{VeryPositive} ones?''. In the rest, we show the effect of the features derived from the word embeddings clusters in the fine-grained classification problem, which was also part of the SemEval-2016 ``Sentiment Analysis in Twitter'' task \cite{NakovRRSS16}. 

\noindent\textbf{Learning Algorithm} To perform the quantification task, we rely on a \textit{classify and count} approach, which was shown effective in a related binary quantification problem \cite{BalikasA16}. The idea is  that given a set of instances on a particular subject, one first classifies the instances and then aggregates the counts. 
To this end, we use the same feature representation steps and data with the ones used for fine grained classification (Section 3.2). Note that the data of the task are associated with subjects (described in full detail at \cite{NakovRRSS16}), and, hence, quantification is performed for the tweets of a subject. For each of the five categories, the output of the approach is a 5-dimensional vector with the estimated prevalence of the categories. 

The evaluation measure for the problem is the Earth Movers Distance (EMD) \cite{rubner2000earth}. EMD is a measure of error, hence lower values are better. It assumes ordered categories, which in our problem is naturally defined. Further assuming that the distance of consecutive categories  (\textit{e.g.,} Positive and VeryPositive) is 1, the measure is calculated by: 

\begin{small}
	\[
	EMD(p, \hat{p}) = \sum\limits_{j=1}^{|C|-1}| \sum\limits_{i=1}^j \hat{p}(c_i)-\sum\limits_{i=1}^j p(c_i)|
	\]
\end{small}

\noindent where $|C|$ is number of categories (five in our case) and $\hat{p}(c_i)$ and $p(c_i)$   are the true and predicted prevalence respectively \cite{EsuliS10}. 

\noindent\textbf{Results} Table \ref{tbl:quant_scores} presents the results of augmenting the feature set with the proposed features. We use Logistic Regression as a base classifier for the classify and count approach. Notice the positive impact of the features in the performance in the task. Adding the features derived from clustering the embeddings consistently  improves the performance. Interestingly, the best performance ($0.219$) is achieved using the out-of-domain vectors, as in the NER classification task. Also, notice how the approach improves over the state-of-the-art performance in the challenge ($0.243$) \cite{NakovRRSS16}, held by the method of \cite{MartinoG016}. The improvement over the method of \cite{MartinoG016} however, does not necessarily mean that classify and count performs better in the task. It implies that the feature set we used is richer, that in turn highlights the value of robust feature extraction mechanisms which is the subject of this paper.    

\begin{table}\scriptsize\centering\setlength{\tabcolsep}{5pt}
	\begin{tabular}{l ccccc}
		\toprule
		clusters& 100 &250 &500&1000 &2000\\
		\midrule
		no clusters & \multicolumn{5}{c}{0.228 [0.243]} \\
		
		skipgram$_{40,w5}$ & 0.223 & 0.223 & \textbf{0.221} & \textbf{0.221} & 0.222 \\
		skipgram$_{100,w5}$ & 0.223 & 0.224 & \textbf{0.221} & 0.223 & 0.222 \\
		skipgram$_{100,w10}$ &  0.225 & 0.223 & \textbf{0.220} & 0.221 & 0.222    \\

		cbow$_{40, w5}$ &  0.227 & \textbf{0.221} & 0.228 & 0.222 & 0.223    \\
		cbow$_{100,w5}$ & 0.225 & 0.225 & 0.227 & \textbf{0.224} & 0.225 \\
		cbow$_{100,w10}$ & 0.226 & 0.226 & 0.231 & \textbf{0.224} & \textbf{0.224} \\
		
		glove$_{40,w5}$    &  0.222 & 0.222 & \textbf{0.220} & 0.221 & 0.222    \\
		glove$_{100,w5}$   &  0.223 & 0.221 & 0.221 & \textbf{0.220} & 0.221  \\
		glove$_{100,w10}$  &  \textbf{0.220} & 0.222 & 0.224 & 0.224 & 0.223    \\\midrule
		
		glove$_{50, wiki}$& 0.222 & 0.221 & \textbf{0.219} & 0.220 & 0.221\\
		glove$_{100,wiki}$& 0.221 & 0.221 & 0.222 & \textbf{0.220} & 0.222 \\
		
		\bottomrule
	\end{tabular}
	\caption{Earth Movers Distance for fine-grained sentiment quantification across different types of word embeddings and number of clusters. The score in brackets denotes the best performance achieved in the challenge.  }\label{tbl:quant_scores}
\end{table}

\section{Conclusion}
We have shown empirically the effectiveness of incorporating cluster membership features in the feature extraction pipeline of Named-Entity recognition, sentiment classification and quantification tasks.
Our results strongly suggest that incorporating cluster membership features benefit the performance in the tasks. The fact that the performance improvements are consistent in the four tasks we investigated, further highlights their usefulness, both for practitioners and researchers. 

Although our study does not identify a clear winner with respect to the type of word vectors (skipgram, cbow, or GloVe), our findings suggest that one should first try skip-gram embeddings of low dimensionality ($D=40$) and high number of clusters (\textit{e.g.,} $K\in\{500,1000,2000\}$) as the results obtained using these settings are consistently competitive. Our results also suggest that using out-of-domain data, like Wikipedia articles in this case, to construct the word embeddings is a good practice, as the results we obtained with these vectors are also competitive. The positive of out-of-domain embeddings and their combination with in-domain ones remains to be further studied.

\bibliography{references}

\begin{thebibliography}{}
\expandafter\ifx\csname natexlab\endcsname\relax\def\natexlab#1{#1}\fi

\bibitem[{Arthur and Vassilvitskii(2007)}]{arthur2007k}
David Arthur and Sergei Vassilvitskii. 2007.
\newblock k-means++: The advantages of careful seeding.
\newblock In {\em ACM-SIAM@Discrete algorithms\/}. pages 1027--1035.

\bibitem[{Baccianella et~al.(2009)Baccianella, Esuli, and
  Sebastiani}]{BaccianellaES09}
Stefano Baccianella, Andrea Esuli, and Fabrizio Sebastiani. 2009.
\newblock Evaluation measures for ordinal regression.
\newblock In {\em International Conference on Intelligent Systems Design and
  Applications\/}. pages 283--287.

\bibitem[{Balikas and Amini(2016)}]{BalikasA16}
Georgios Balikas and Massih{-}Reza Amini. 2016.
\newblock Twise at semeval-2016 task 4: Twitter sentiment classification.
\newblock In {\em Proceedings of the 10th International Workshop on Semantic
  Evaluation, SemEval@NAACL-HLT 2016, San Diego, CA, USA, June 16-17, 2016\/}.
  pages 85--91.

\bibitem[{Brown et~al.(1992)Brown, deSouza, Mercer, Pietra, and Lai}]{Brown92}
Peter~F. Brown, Peter~V. deSouza, Robert~L. Mercer, Vincent J.~Della Pietra,
  and Jenifer~C. Lai. 1992.
\newblock Class-based n-gram models of natural language.
\newblock {\em Computational Linguistics\/} 18:467--479.

\bibitem[{{Daum{\'{e}} III} et~al.(2014){Daum{\'{e}} III}, Langford, and
  Ross}]{Daume14}
Hal {Daum{\'{e}} III}, John Langford, and St{\'{e}}phane Ross. 2014.
\newblock Efficient programmable learning to search.
\newblock {\em CoRR\/} abs/1406.1837.

\bibitem[{Dey et~al.(2016)Dey, Shrivastava, and Kaushik}]{DeySK16}
Kuntal Dey, Ritvik Shrivastava, and Saroj Kaushik. 2016.
\newblock A paraphrase and semantic similarity detection system for user
  generated short-text content on microblogs.
\newblock In {\em COLING\/}. pages 2880--2890.

\bibitem[{Esuli and Sebastiani(2010)}]{EsuliS10}
Andrea Esuli and Fabrizio Sebastiani. 2010.
\newblock Sentiment quantification.
\newblock {\em {IEEE} Intelligent Systems\/} 25(4):72--75.

\bibitem[{Goldberg(2015)}]{Goldberg15c}
Yoav Goldberg. 2015.
\newblock \href{http://arxiv.org/abs/1510.00726}{A primer on neural network
  models for natural language processing}.
\newblock {\em CoRR\/} abs/1510.00726.
\newblock
  \href{http://arxiv.org/abs/1510.00726}{http://arxiv.org/abs/1510.00726}.

\bibitem[{Guo et~al.(2014)Guo, Che, Wang, and Liu}]{guo2014revisiting}
Jiang Guo, Wanxiang Che, Haifeng Wang, and Ting Liu. 2014.
\newblock Revisiting embedding features for simple semi-supervised learning.
\newblock In {\em EMNLP\/}. pages 110--120.

\bibitem[{Hee et~al.(2016)Hee, Lefever, and Hoste}]{HeeLH16}
Cynthia~Van Hee, Els Lefever, and V{\'{e}}ronique Hoste. 2016.
\newblock Monday mornings are my fave : {)} {\#}not exploring the automatic
  recognition of irony in english tweets.
\newblock In {\em COLING\/}. pages 2730--2739.

\bibitem[{Hu and Liu(2004)}]{HuL04}
Minqing Hu and Bing Liu. 2004.
\newblock \href{https://doi.org/10.1145/1014052.1014073}{Mining and summarizing
  customer reviews}.
\newblock In {\em SIGKDD\/}. pages 168--177.
\newblock
  \href{https://doi.org/10.1145/1014052.1014073}{https://doi.org/10.1145/1014052.1014073}.

\bibitem[{Kiritchenko et~al.(2014)Kiritchenko, Zhu, and
  Mohammad}]{KiritchenkoZM14}
Svetlana Kiritchenko, Xiaodan Zhu, and Saif~M. Mohammad. 2014.
\newblock Sentiment analysis of short informal texts.
\newblock {\em JAIR\/} 50:723--762.

\bibitem[{Martino et~al.(2016)Martino, Gao, and Sebastiani}]{MartinoG016}
Giovanni Da~San Martino, Wei Gao, and Fabrizio Sebastiani. 2016.
\newblock Ordinal text quantification.
\newblock In {\em SIGIR\/}. pages 937--940.

\bibitem[{Mikolov et~al.(2013)Mikolov, Chen, Corrado, and Dean}]{abs-1301-3781}
Tomas Mikolov, Kai Chen, Greg Corrado, and Jeffrey Dean. 2013.
\newblock Efficient estimation of word representations in vector space.
\newblock {\em CoRR\/} abs/1301.3781.

\bibitem[{Nakov et~al.(2016)Nakov, Ritter, Rosenthal, Sebastiani, and
  Stoyanov}]{NakovRRSS16}
Preslav Nakov, Alan Ritter, Sara Rosenthal, Fabrizio Sebastiani, and Veselin
  Stoyanov. 2016.
\newblock \href{http://aclweb.org/anthology/S/S16/S16-1001.pdf}{Semeval-2016
  task 4: Sentiment analysis in twitter}.
\newblock In {\em SemEval@NAACL-HLT 2016\/}. pages 1--18.
\newblock
  \href{http://aclweb.org/anthology/S/S16/S16-1001.pdf}{http://aclweb.org/anthology/S/S16/S16-1001.pdf}.

\bibitem[{Owoputi et~al.(2013)Owoputi, O'Connor, Dyer, Gimpel, Schneider, and
  Smith}]{Owoputi13}
Olutobi Owoputi, Brendan O'Connor, Chris Dyer, Kevin Gimpel, Nathan Schneider,
  and Noah~A. Smith. 2013.
\newblock Improved part-of-speech tagging for online conversational text with
  word clusters.
\newblock In {\em NAACL\/}. pages 380--390.

\bibitem[{Partalas et~al.(2016)Partalas, Lopez, Derbas, and
  Kalitvianski}]{Partalas16c}
Ioannis Partalas, C\'{e}dric Lopez, Nadia Derbas, and Ruslan Kalitvianski.
  2016.
\newblock Learning to search for recognizing named entities in twitter.
\newblock In {\em W-NUT, Coling\/}.

\bibitem[{Pennington et~al.(2014)Pennington, Socher, and Manning}]{glove}
Jeffrey Pennington, Richard Socher, and Christopher~D. Manning. 2014.
\newblock Glove: Global vectors for word representation.
\newblock In {\em EMNLP\/}. pages 1532--1543.

\bibitem[{Ritter et~al.(2011)Ritter, Clark, Mausam, and Etzioni}]{Ritter2011}
Alan Ritter, Sam Clark, Mausam, and Oren Etzioni. 2011.
\newblock Named entity recognition in tweets: An experimental study.
\newblock In {\em Proceedings of the Conference on Empirical Methods in Natural
  Language Processing\/}. EMNLP '11, pages 1524--1534.

\bibitem[{Rubner et~al.(2000)Rubner, Tomasi, and Guibas}]{rubner2000earth}
Yossi Rubner, Carlo Tomasi, and Leonidas~J Guibas. 2000.
\newblock The earth mover's distance as a metric for image retrieval.
\newblock {\em International journal of computer vision\/} 40(2):99--121.

\bibitem[{Salton et~al.(1975)Salton, Wong, and Yang}]{SaltonWY75}
Gerard Salton, A.~Wong, and C.~S. Yang. 1975.
\newblock \href{https://doi.org/10.1145/361219.361220}{A vector space model for
  automatic indexing}.
\newblock {\em Commun. {ACM}\/} 18(11):613--620.
\newblock
  \href{https://doi.org/10.1145/361219.361220}{https://doi.org/10.1145/361219.361220}.

\bibitem[{Strauss et~al.(2016)Strauss, Toma, Ritter, de~Marneffe, and
  Xu}]{Strauss16}
Benjamin Strauss, Bethany~E. Toma, Alan Ritter, Marie~Catherine de~Marneffe,
  and Wei Xu. 2016.
\newblock Results of the wnut16 named entity recognition shared task.
\newblock In {\em WNUT@COLING\/}. Osaka, Japan.

\bibitem[{Turian et~al.(2010)Turian, Ratinov, and Bengio}]{turian2010word}
Joseph Turian, Lev Ratinov, and Yoshua Bengio. 2010.
\newblock Word representations: a simple and general method for semi-supervised
  learning.
\newblock In {\em ACL\/}. pages 384--394.

\bibitem[{Wang and Manning(2013)}]{wang2013effect}
Mengqiu Wang and Christopher~D Manning. 2013.
\newblock Effect of non-linear deep architecture in sequence labeling.
\newblock In {\em IJCNLP\/}. pages 1285--1291.

\bibitem[{Wilson et~al.(2005)Wilson, Wiebe, and Hoffmann}]{WilsonWH05}
Theresa Wilson, Janyce Wiebe, and Paul Hoffmann. 2005.
\newblock Recognizing contextual polarity in phrase-level sentiment analysis.
\newblock In {\em EMNLP\/}.

\end{thebibliography}
\bibliographystyle{acl_natbib}

\end{document}